\documentclass[11pt]{article}

\usepackage[final]{acl}

\usepackage{times}
\usepackage{latexsym}

\usepackage{tcolorbox}
\tcbuselibrary{skins, breakable}

\usepackage{fancyvrb}

\usepackage[T1]{fontenc}

\usepackage[utf8]{inputenc}

\usepackage{microtype}

\usepackage{inconsolata}
\usepackage{amsmath,amssymb}
\usepackage{graphicx}
\usepackage{adjustbox}
\usepackage{caption}
\usepackage{siunitx}
\usepackage{booktabs}
\usepackage{makecell}
\usepackage{graphicx}

\usepackage{booktabs}   
\usepackage{multirow}   
\usepackage{makecell}   
\usepackage{array}
\usepackage{siunitx}
\usepackage{subcaption}

\usepackage{xcolor}
\newif\ifdraft\draftfalse

\title{Beyond Scores: Understanding LLM-as-a-Judge Mechanisms in Summarization Evaluation}

\author{
    Himil Vasava, Ming Jiang \\
    University of Wisconsin-Madison \\
    \texttt{\{vasava, ming.jiang\}@wisc.edu}
    }

\begin{document}
\maketitle

\begin{abstract}
LLM-based evaluators of natural language generation (NLG) quality are widely deployed as scoring tools and as automated training signals, yet the internal procedure by which they assign a rating remains poorly understood. We investigate this procedure mechanistically through an eight-attack perturbation taxonomy across the Readability and Adequacy dimensions of NLG quality, a generation pipeline that produces paired clean and corrupt summaries with controlled error intensity and explicit token-level modification maps, and a four-experiment battery of causal tracing, logit-lens vocabulary projection, and attention-head knockout applied to Themis (Llama-3-8B) and Prometheus (Mistral-7B). Both evaluators implement a structured, coherent evaluation pipeline operating in two stages: below layer~$15$, attention performs local error comparison and routes the result to the final input position; above it, the MLP cascade integrates the signal and writes the rating, with the decision crystallizing in the residual stream at a sharp late layer ($L\!=\!26$ on Themis, $L\!=\!25$ on Prometheus). Furthermore, a base-model control at the same scale (Llama-3-8B) reproduces the routing architecture and crystallization but not the stage separation, isolating the two mechanisms that fine-tuning specifically installs, suppression of below-L15 MLP contribution at the last position and a two-layer advance of the crystallization depth, indicating that fine-tuning sculpts an existing substrate rather than building the pipeline from scratch. We release the source code and data at \url{https://github.com/himil-v/judge-mech}.
\end{abstract}

\section{Introduction}
\label{sec:intro}


Automatic evaluation using large language models (LLMs) has become widely adopted in natural language generation (NLG): dedicated evaluator models such as Themis~\citep{hu2024themis} and Prometheus~\citep{kim2024prometheus}, alongside prompting-based approaches including G-Eval~\citep{liu2023g} and MT-Bench~\citep{zheng2023judging}, are now routinely applied to tasks ranging from summarization faithfulness to dialogue coherence assessment and increasingly serve as automated signals for preference labeling and reward modeling. Existing examination of these LLM-as-a-judge systems has focused largely on behavioral analysis: measuring agreement with human ratings~\citep{fabbri2021summeval,liu2023g}, cataloging failure modes such as criterion confusion~\citep{hu2024llm}, and quantifying positional and length biases~\citep{zheng2023judging}. This paradigm characterizes \emph{what} LLM-judges score but leaves \emph{how} the rating is computed inside the model largely unexamined, limiting both evaluator reliability and our ability to anticipate failure cases.

\begin{figure*}[t]
  \centering
  \includegraphics[width=0.8\textwidth]{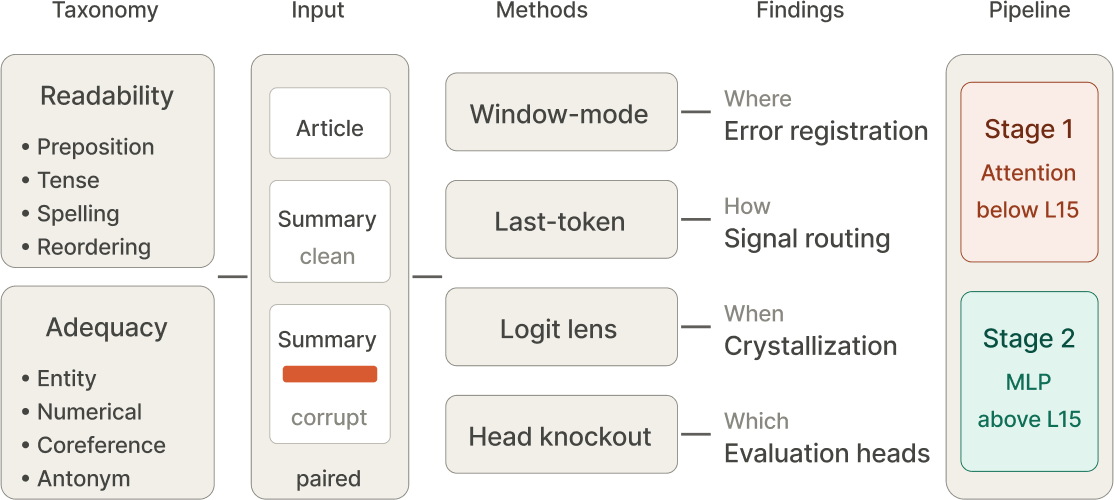}
  \caption{Overview of our analysis. We design an eight-attack \textbf{perturbation taxonomy} across the Readability and Adequacy dimensions of NLG quality, and generate paired clean/corrupt \textbf{summaries} of an article with controlled error intensity and explicit token-level modification maps. Four \textbf{mechanistic methods} - window-mode and last-token causal tracing, logit lens, and attention-head knockout, each reveal a different aspect of how the rating is computed: \emph{where} errors are registered, \emph{how} the signal routes to the output position, \emph{when} the decision crystallizes in the residual stream, and \emph{which} heads implement the verdict.}
\label{fig:teaser}
\end{figure*}

We address this gap through a mechanistic study, asking: \emph{Does the model identify the specific quality defect, or does it produce a rating consistent with training-distribution patterns without engaging the underlying error?} We focus on text summarization as a case study because it is one of the most extensively studied NLG evaluation tasks, providing mature perturbation methods and well-established benchmark datasets. Moreover, its structured article-summary format naturally facilitates localized perturbation interventions. 

One key challenge in directly applying causal tracing to LLM-judges is that these methods require paired clean and corrupted inputs with precisely tracked token-level modifications and corresponding rating shifts. However, existing adversarial summarization benchmarks typically introduce errors at the sentence or document level, making it difficult to isolate and trace their effects at the token level. To address this limitation, we develop a new perturbation taxonomy consisting of eight localized attack types that span the \emph{Readability} and \emph{Adequacy} criteria of NLG quality, along with a generation pipeline that produces paired clean/corrupt summaries with controlled error intensity and explicit token-level modification maps. Using this framework, we investigate the internal mechanisms of two state-of-the-art NLG LLM judges, each fine-tuned from a different model family, to uncover how they internally process and assign scores for summary quality. Our key findings include:

\begin{enumerate}
    \item LLM-based evaluator models exhibit structured, coherent information propagation during evaluation, rather than an ad hoc mapping from surface features to rating tokens. Specifically, the attention module in the lower layers ($<15$ layer) primarily performs error identification and signal routing, while the upper-layer MLP integrates these routed signals to produce the final rating, with the decision crystallizing in the residual stream at a sharp late layer ($\sim 25 - 26$ layer). 
    \item The process of error identification in attention varies systematically across error types: readability-related errors spread across adjacent syntactic context, while adequacy attacks remain concentrated on the perturbed tokens. This indicates that LLM-judges deploy distinct attention strategies depending on the criterion, suggesting that criterion-specific attention supervision could improve evaluator robustness. 
    \item The way the final rating is written into the residual stream differs between the two evaluators studied: Themis commits to its rating in a single dominant step at the top layer, while Prometheus spreads this write across mid-to-late layers and applies an opposite-direction correction at the final layer.
    \item Comparing the fine-tuned evaluators to their base architecture 
    (Llama-3-8B) reveals that fine-tuning does not build the 
    evaluation pipeline from scratch: the base model already exhibits the 
    routing architecture and sharp late-layer crystallization property (at 
    $L\!=\!28$), and its above L15 MLP magnitude matches Themis's. Judge 
    fine-tuning installs two specific modifications on this general 
    substrate, suppression of below L15 MLP contribution at the last 
    position (producing the stage-1/stage-2 separation) and a two-layer 
    advance of the crystallization depth (from $L\!=\!28$ to $L\!=\!26$).
\end{enumerate}

To the best of our knowledge, this is the first study to systematically uncover the internal mechanisms underlying LLM-based NLG evaluation, with an emphasis on text summarization. The resulting insights hope to benefit the design of more robust and interpretable LLM-based evaluators.


\section{Related Work}
\label{sec:related}

\paragraph{LLM-based NLG evaluation.} Dedicated evaluator models~\citep{hu2024themis,kim2024prometheus} and prompting-based scoring frameworks~\citep{liu2023g,zheng2023judging} have become standard tools for assessing NLG quality, and prior work has examined evaluator failure modes including criterion confusion and prompt sensitivity~\citep{hu2024llm,zheng2023judging}. Established factual-consistency and faithfulness benchmarks~\citep{fabbri2021summeval,kryscinski2020evaluating,pagnoni2021understanding,honovich2022true,dziri2022faithdial} provide the empirical foundation for evaluator development but remain behavioral in scope, characterizing what LLM-judges score rather than how they arrive at those scores. The present work complements this literature by examining the evaluator's internal procedure: rather than measuring agreement with human judgments, we analyze how the rating is computed inside the model.

\paragraph{Mechanistic interpretability of transformer language models.} A growing body of work localizes computation inside transformer models through activation patching and causal mediation analysis~\citep{vig2020causal,meng2022locating}, vocabulary-space projection of intermediate residual streams~\citep{nostalgebraist2020}, and circuit-level identification of attention-head structure~\citep{wang2023interpretability}. These methods have been applied to factual recall~\citep{meng2022locating}, indirect-object identification~\citep{wang2023interpretability}, and other narrowly-scoped behaviors, but to our knowledge LLM-based NLG evaluation has not previously been the subject of mechanistic analysis. We extend these methods to the evaluator setting, applying them in combination across two evaluator models drawn from distinct base-model families.

\begin{table}[t]
\centering
\small
\resizebox{0.48\textwidth}{!}{%
\begin{tabular}{ll}
\toprule
\textbf{Final Node} & \textbf{Source Perturbations} \\
\midrule
\multicolumn{2}{l}{\textit{Readability}} \\
Preposition Mismatch & grammatical error~\citep{pagnoni2021understanding}, \\
                     & subject-verb disagreement~\citep{ribeiro2020beyond} \\
Tense Mismatch       & incorrect verb form~\citep{pagnoni2021understanding}, \\
                     & grammatical error~\citep{pagnoni2021understanding} \\
Spelling             & character-level noise~\citep{belinkov2017synthetic} \\
Sequential Reordering & jumbling word order~\citep{ribeiro2020beyond}, \\
                     & sentence exchange~\citep{kryscinski2020evaluating} \\
\midrule
\multicolumn{2}{l}{\textit{Adequacy}} \\
Entity Swap          & entity swap~\citep{kryscinski2020evaluating}, \\
                     & entity error~\citep{pagnoni2021understanding}, \\
                     & perturb names/nouns~\citep{ribeiro2020beyond} \\
Numerical/Date Swap  & number swap~\citep{kryscinski2020evaluating}, \\
                     & change numbers~\citep{ribeiro2020beyond} \\
Coreference Mismatch & coreference error~\citep{pagnoni2021understanding}, \\
                     & pronoun swap~\citep{kryscinski2020evaluating} \\
Antonym \& Negation  & sentence negation~\citep{kryscinski2020evaluating}, \\
                     & negation/antonyms~\citep{ribeiro2020beyond} \\
\bottomrule
\end{tabular}
}
\caption{Consolidation mapping showing how perturbation types from prior NLG-evaluation benchmarks are unified into the eight nodes of our taxonomy across the Readability and Adequacy dimensions.}

\label{tab:taxonomy-consolidation}
\end{table}

\paragraph{Perturbation-based evaluation of NLG systems.} Adversarial perturbation has long been used for diagnostic evaluation of NLG and natural-language inference systems, ranging from character-level noise~\citep{belinkov2017synthetic} and behavioral testing~\citep{ribeiro2020beyond} to factuality-focused perturbations of generated summaries~\citep{kryscinski2020evaluating,pagnoni2021understanding,li2023halueval,min2023factscore}. These resources have established that perturbations can isolate specific failure modes, but they are designed for behavioral benchmarking and do not preserve the position-level metadata required for mechanistic analysis. Our perturbation taxonomy and generation pipeline build on this prior methodology while introducing the additional controls necessary for mechanistic study.
\section{Data Perturbation}
\label{sec:taxonomy}

Existing adversarial datasets for NLG evaluation primarily focus on behavioral benchmarking and typically introduce errors at the sentence or document level. Such coarse-grained perturbations are insufficient for mechanistic analysis, which requires paired clean and corrupted inputs with explicit token-level modification maps. To enable such analysis, we design a perturbation framework with two key controls: \emph{controlled intensity}, which fixes the number of perturbed tokens in each sample through a parameter $k$, and \emph{spatial localization}, which records the positional indices of all perturbed tokens, allowing causal tracing to anchor its analysis to known perturbation sites.


\paragraph{Selection criteria.} Three considerations narrow the space of 
possible perturbations to the eight error types we adopt. First, 
\emph{grounding in prior work}: every node consolidates categories 
established in published NLG-evaluation benchmarks 
(Table~\ref{tab:taxonomy-consolidation}), not invented for this study. 
Second, \emph{method compatibility}: causal tracing requires a 
token-localizable perturbed span and a single, isolable failure mode, 
which excludes stylistic alterations (multi-token, semantically 
entangled) and coarse noise injections that obscure intermediate 
representations. Third, \emph{criterion coverage}: we retain four attacks 
per quality axis (Readability and Adequacy, following 
\citep{fabbri2021summeval, honovich2022true, liu2023g}) so that 
criterion-level effects can be compared across balanced categories.

\begin{figure*}[!t]
\centering
\begin{subfigure}{0.75\textwidth}
\centering
\includegraphics[width=\textwidth]{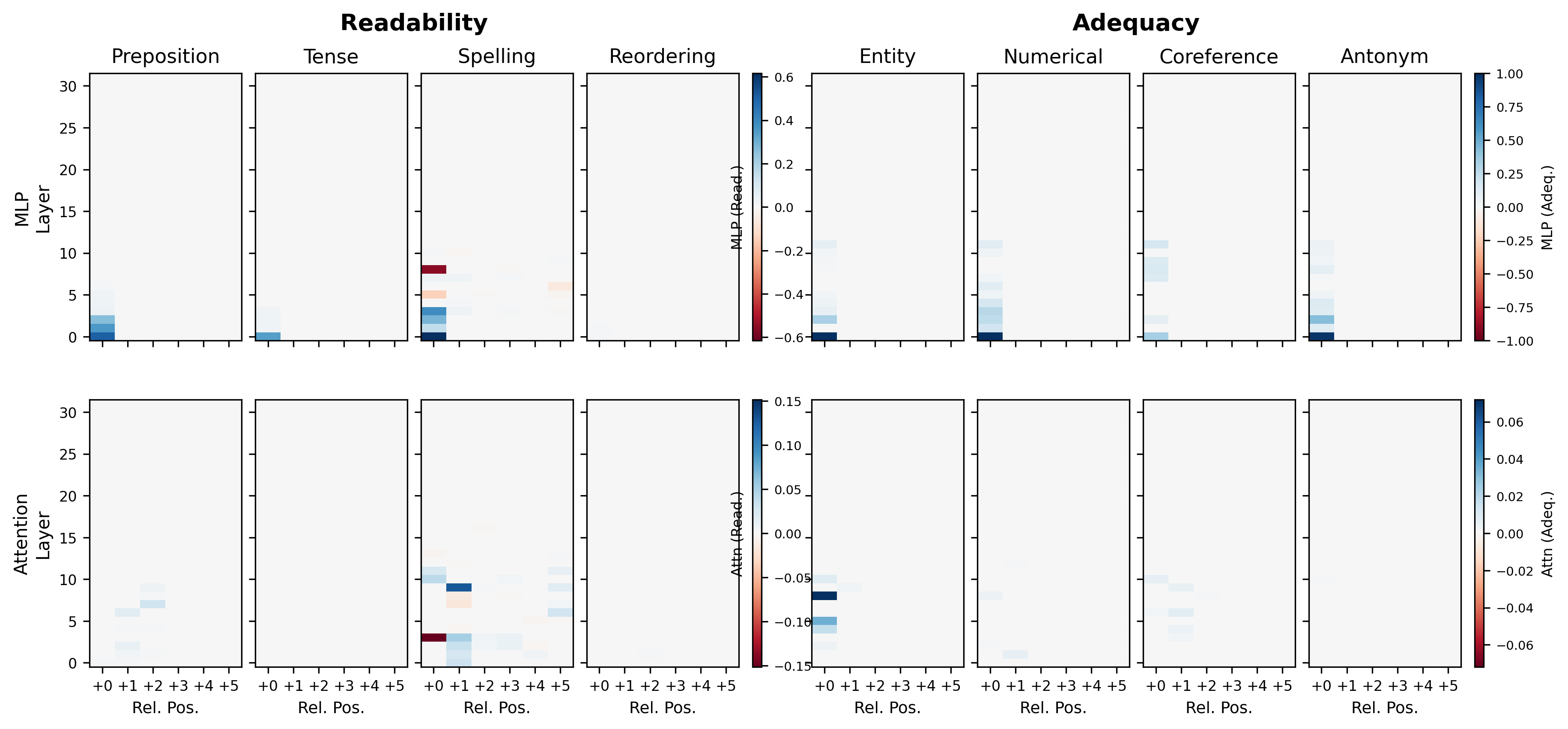}
\caption{Themis (Llama-3-8B)}
\label{fig:window-themis}
\end{subfigure}
\\[6pt]
\begin{subfigure}{0.75\textwidth}
\centering
\includegraphics[width=\textwidth]{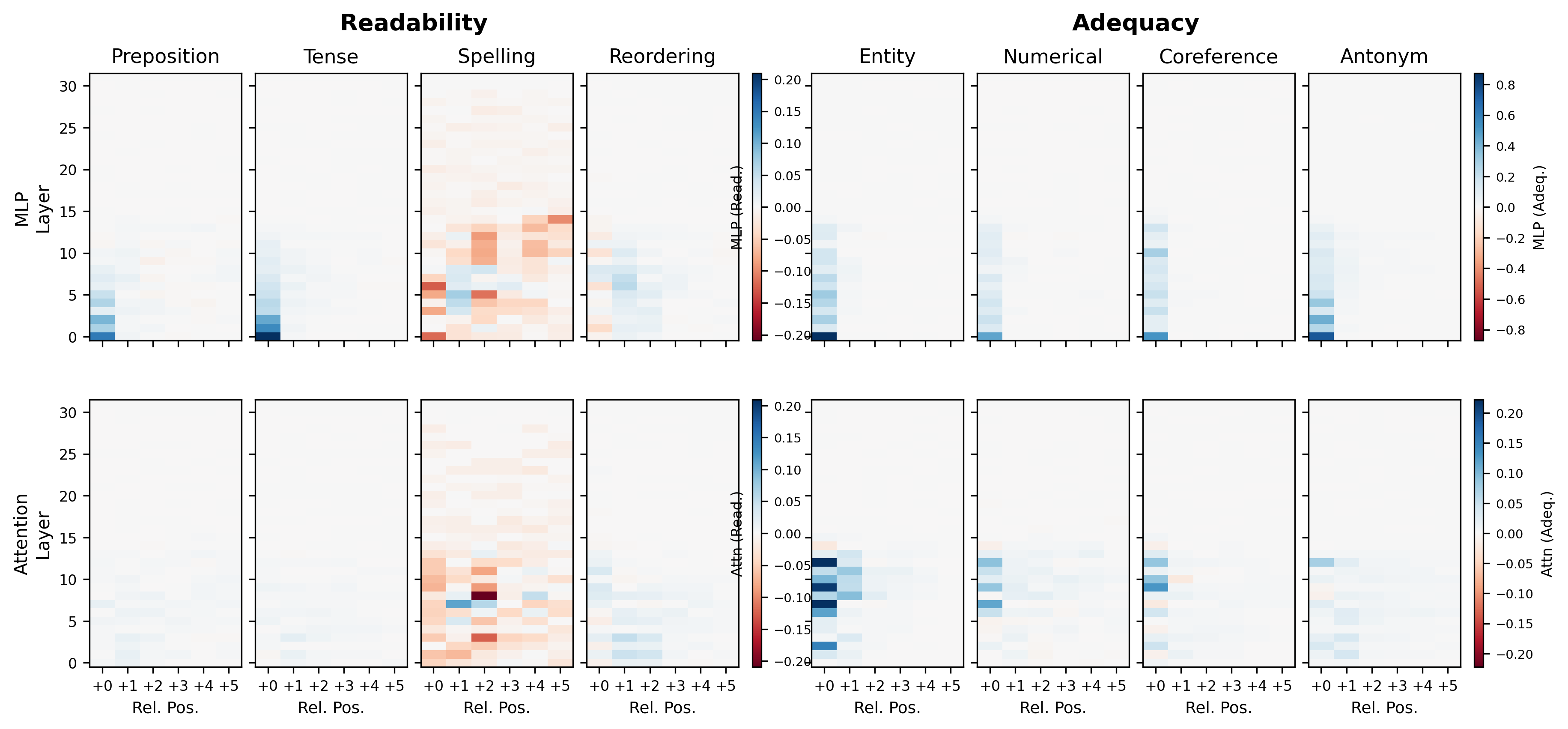}
\caption{Prometheus (Mistral-7B)}
\label{fig:window-prometheus}
\end{subfigure}
\caption{Window-mode causal tracing across all eight attacks on both evaluators: (a) Themis, (b) Prometheus. Top row of each panel: MLP causal effects. Bottom row: attention $o_\text{proj}$ causal effects. Effects are normalized so that $1.0$ recovers clean expected-rating behavior and $0.0$ matches the unmodified corrupt baseline; per-cell aggregates use a $20\%$ trimmed mean over $3$ seeds $\times\ 200$ samples per attack. Four independent color scales per panel preserve within-criterion visibility because effect magnitudes differ by roughly an order of magnitude between Readability and Adequacy.}
\label{fig:window-mode}
\end{figure*}

\paragraph{Perturbation Taxonomy.} Starting from an initial set of over $35$ error 
types drawn from prior perturbation benchmarks 
\citep{kryscinski2020evaluating, pagnoni2021understanding, 
ribeiro2020beyond}, we merge categories targeting the same underlying 
failure mode (e.g., \emph{entity swap}, \emph{entity error}, and 
\emph{perturb names/nouns} $\rightarrow$ \emph{Entity Swap}; 
Table~\ref{tab:taxonomy-consolidation}) and exclude categories 
incompatible with the selection criteria above. The result is eight 
perturbation types. \textbf{Readability Attacks} degrade fluency while 
preserving factual content: \emph{Preposition Mismatch}, \emph{Tense 
Mismatch}, and \emph{Spelling Mistakes} induce localized syntactic or 
orthographic disruptions \citep{ribeiro2020beyond, belinkov2017synthetic, 
hu2024llm}, and \emph{Sequential Reordering} disrupts clause- and 
sentence-level coherence \citep{zheng2023judging}. \textbf{Adequacy 
Attacks} target factual grounding: \emph{Entity Swap} and 
\emph{Numerical/Date Swap} corrupt atomic factual units 
\citep{kryscinski2020evaluating, li2023halueval, min2023factscore}, 
\emph{Coreference Mismatch} alters referential dependencies 
\citep{dziri2022faithdial}, and \emph{Antonym \& Negation} injects 
logical contradictions \citep{bowman2015large, williams2018broad, 
kryscinski2020evaluating}

\paragraph{Automated Perturbation Generation.} We operationalize the taxonomy at scale via an automated generation pipeline. Clean summaries are drawn from CNN/DailyMail \citep{hermann2015teaching}, the same pipeline is applied to XSum \citep{narayan-etal-2018-dont} to produce the cross-domain corpus used in the generalization test. For each perturbation category and intensity $k$, GPT-4o is prompted with category-specific few-shot exemplars and instructed to inject exactly $k$ instances of the targeted error. To eliminate the need for post-hoc text alignment, the generator returns not only the perturbed text but also explicit \texttt{original\_tokens} and \texttt{new\_tokens} arrays, providing a token-level map of the attack. Outputs are constrained to a predefined JSON schema via API-level response formatting. Data quality is verified by a per run manual inspection protocol (Appendix~\ref{app:verification}): approximately 20 clean/corrupt pairs per attack are checked against three criteria, that a perturbation was actually injected, that the returned token maps identify the changed span correctly, and that the perturbation matches its declared type, with typically 19 to 20 of 20 passing.

A filter retains only samples suitable for causal-tracing analysis: those that fail JSON decoding or violate the requested intensity bounds are dropped structurally, and samples where the evaluator's expected rating shifts by less than a threshold between clean and perturbed inputs are excluded from the main analysis because the normalized causal effect is numerically ill-conditioned in that regime. 


\section{LLM-judge Interpretation}

\paragraph{Activation Patching.} We localize where in the network a perturbation's effect is processed by substituting clean activations into a corrupt forward pass at specific (layer, position) sites and measuring how much of the clean rating behavior is recovered~\citep{vig2020causal,meng2022locating}. We trace both MLP sublayers and attention output projections ($o_\text{proj}$), and quantify each intervention with a normalized causal effect
\begin{equation}
\text{effect} = \frac{X_\text{patched} - X_\text{corrupt}}{X_\text{clean} - X_\text{corrupt}},
\end{equation}
where $X$ is an output statistic at the final position, scaled so that $1.0$ recovers clean rating behavior and $0.0$ matches the unmodified corrupt baseline. We use the standard logit difference $z[t_\text{clean}] - z[t_\text{corrupt}]$ for last-token-mode tracing and head knockout; for window-mode tracing, where per-token BPE alignment reduces per-sample yield, we use the expected rating $X = \sum_{r=1}^{5} r \cdot P(r)$ over the renormalized rating-token distribution~\citep{liu2023g}. \emph{Window-mode} tracing restores activations in a six-token window starting at the perturbed token (relative positions $+0$ through $+5$), localizing where in the input the perturbation is processed; filtering, alignment, and aggregation details are in Appendix~\ref{app:tracing-details}. \emph{Last-token-mode} tracing restores activations only at the final input position at each layer, localizing the depth at which the rating decision is assembled.

\paragraph{Logit Lens.} To trace at what depth the model's decision forms, we apply the logit lens~\citep{nostalgebraist2020} to the residual stream at the final input position: at each layer $\ell$, we project the residual through the final layer norm and unembedding to obtain a layer-$\ell$ distribution over the five rating tokens $\{t_1, \ldots, t_5\}$. Sweeping $\ell$ across all layers produces a depth-resolved trajectory for each rating, from which we read both the crystallization depth and the rating ultimately committed to.

\paragraph{Attention Head Knockout.} Attention head knockout identifies which individual heads implement the rating decision~\citep{wang2023interpretability}. For each of the $32 \times 32 = 1024$ heads in each model, we zero out that head's output during the forward pass on corrupt inputs and compute the resulting normalized causal effect (Equation~1) at the final position. Positive (red) effects identify \emph{disruptive} heads that drive the corrupt rating - ablation releases the clean signal, while negative (blue) effects identify \emph{suppressive} heads that defend the clean rating. We report per-attack heatmaps and inspect criterion-level patterns by comparing the four panels within each criterion group.

\section{Experiments}
\label{sec:methods}

\subsection{Experimental Setup}
\label{sec:setup}
We study two open-source LLM-judges from different base-model families: Themis~\citep{hu2024themis}, built on Llama-3-8B, and Prometheus-7B-v2.0~\citep{kim2024prometheus}, built on Mistral-7B. Both have $32$ transformer layers and $32$ attention heads per layer. For each of the eight perturbations from §\ref{sec:taxonomy}, we draw $200$ clean/corrupt summary pairs from our CNN/DailyMail-derived corpus and score both texts with the target evaluator using a $1$-$5$ NLG-evaluation prompt (Appendix~\ref{app:prompts}); Prometheus additionally uses a few-shot variant to place the rating token at the first generation position. Following the behavioral filter (§\ref{sec:taxonomy}), we retain only pairs whose rating changes after perturbation. All results are averaged over three seeds and $200$ samples per attack.


\begin{figure}[!t]
  \includegraphics[width=0.88\columnwidth]{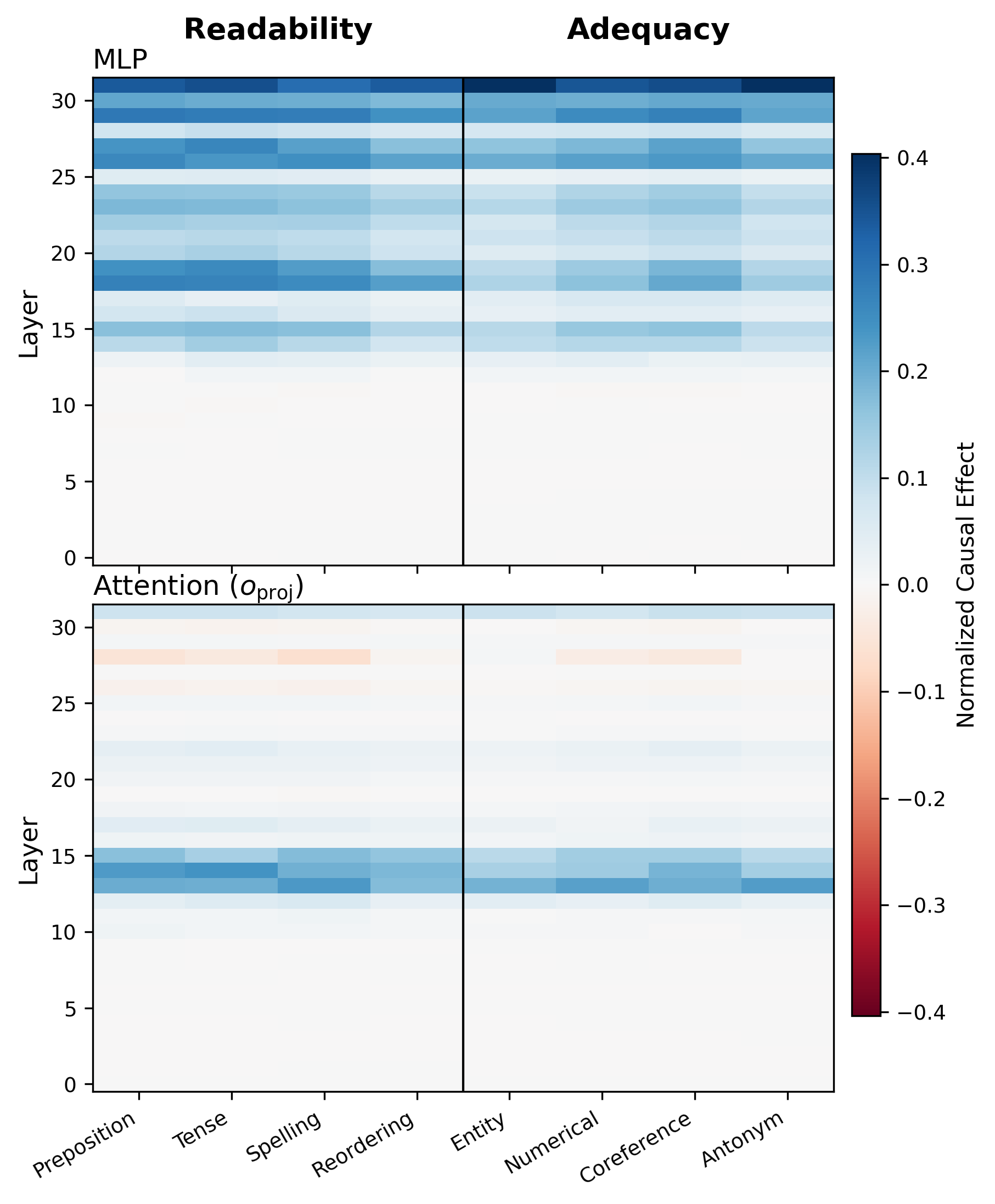}
  \caption{Last Token mode causal tracing on Themis across all eight attacks. Top row: MLP causal effects; bottom row: Attention $o_{\text{proj}}$ causal effects.}
  \label{fig:themis-last}
\end{figure}

\begin{figure}[!t]
  \includegraphics[width=0.88\columnwidth]{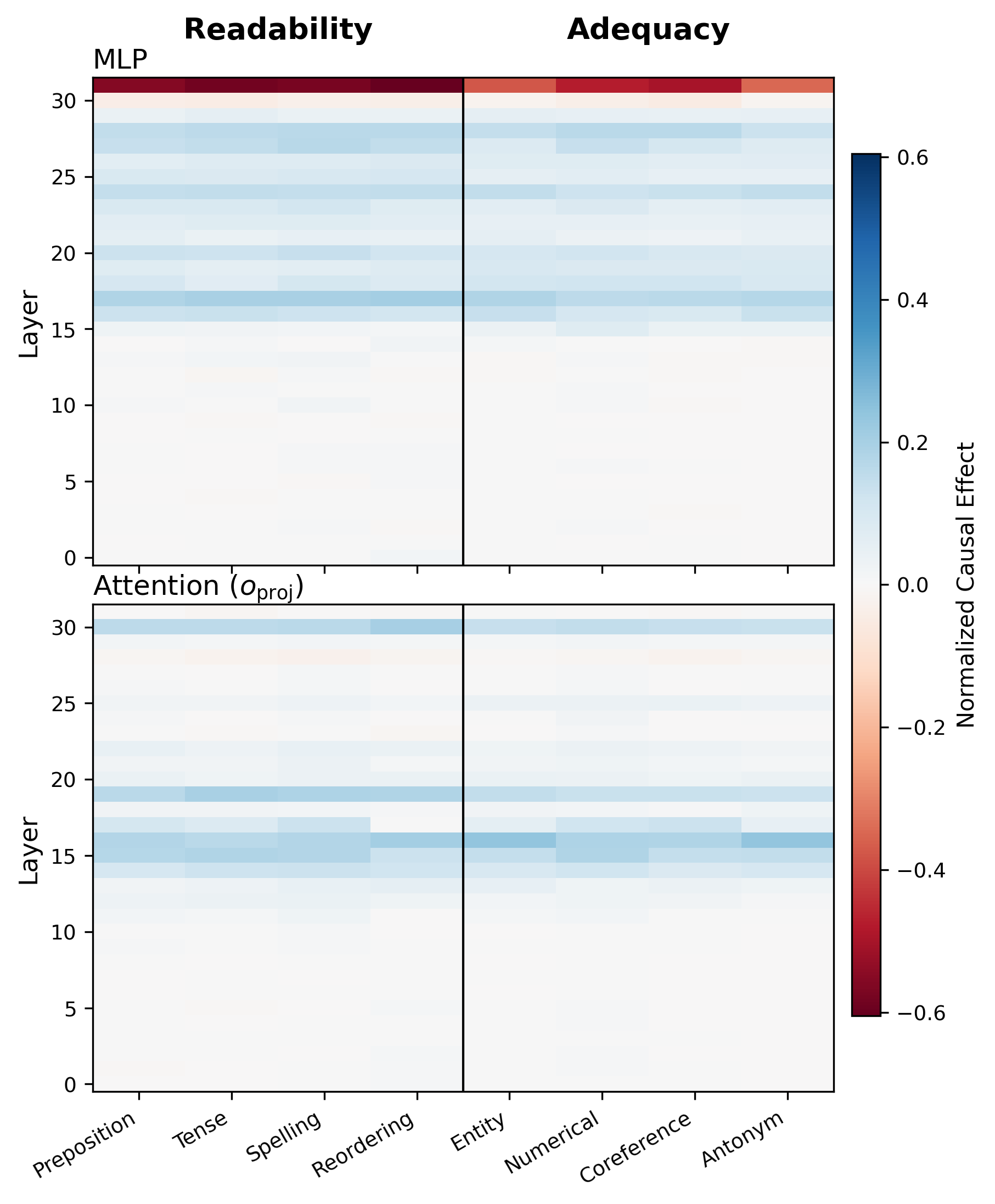}
  \caption{Last Token mode causal tracing on Prometheus across all eight attacks. Top row: MLP causal effects; bottom row: Attention $o_{\text{proj}}$ causal effects.}
  \label{fig:prometheus-last}
\end{figure}


\begin{figure*}[t]
\centering
\begin{subfigure}{0.68\textwidth}
\centering
\includegraphics[width=\textwidth]{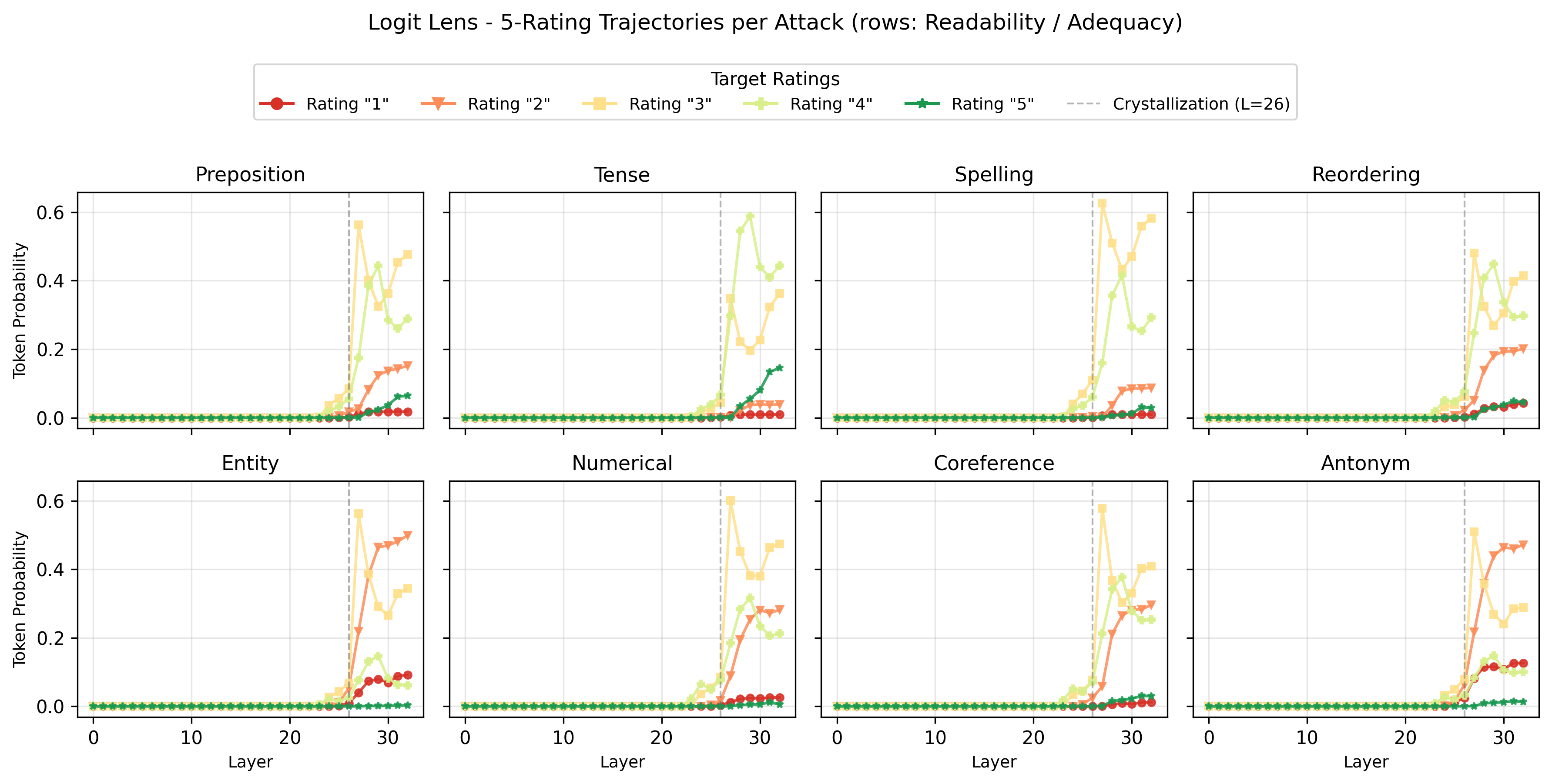}
\caption{Themis (Llama-3-8B), crystallization at $L\!=\!26$}
\label{fig:logit-lens-themis}
\end{subfigure}
\\[6pt]
\begin{subfigure}{0.68\textwidth}
\centering
\includegraphics[width=\textwidth]{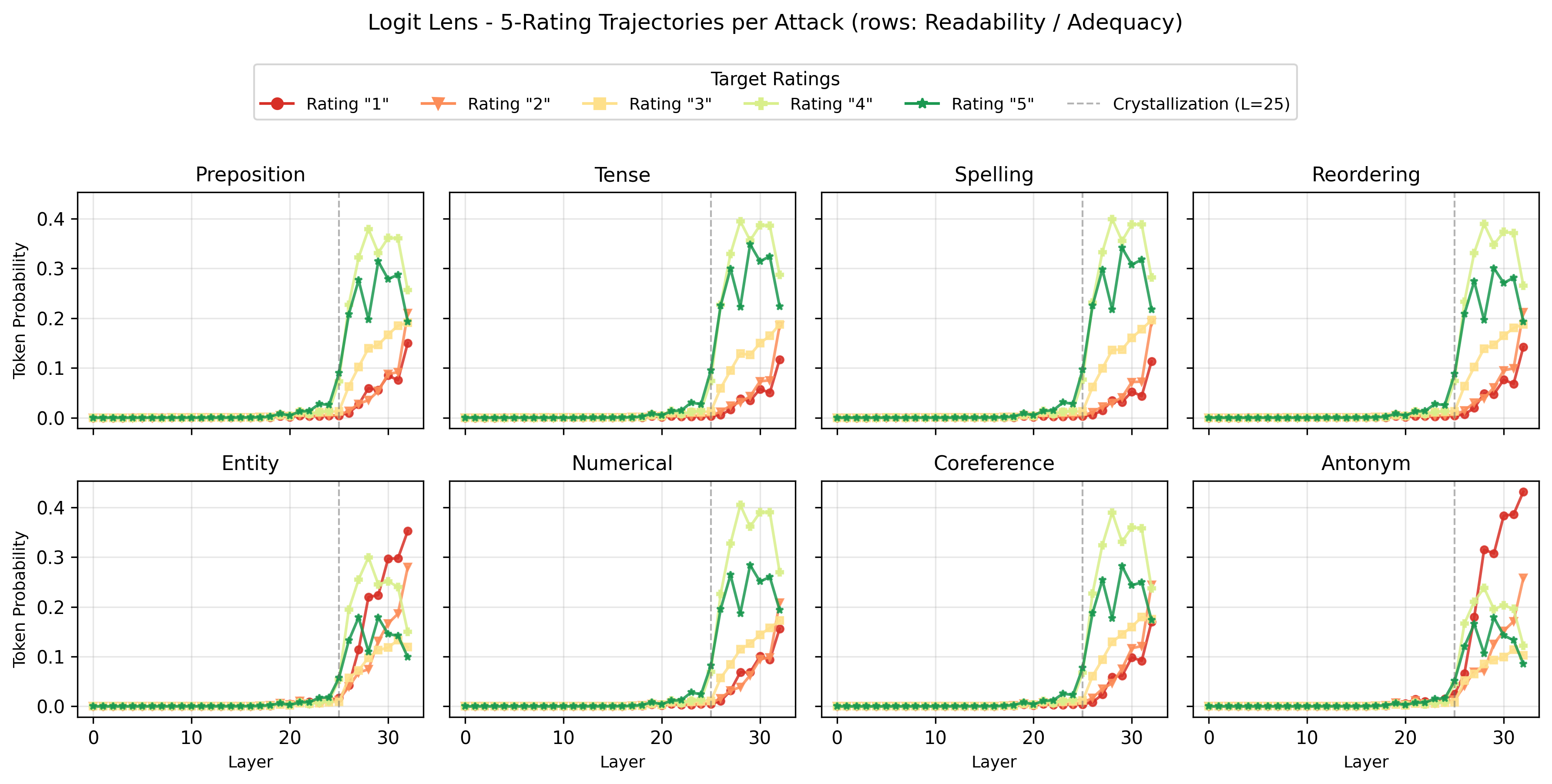}
\caption{Prometheus (Mistral-7B), crystallization at $L\!=\!25$}
\label{fig:logit-lens-prometheus}
\end{subfigure}
\caption{Logit-lens trajectories for each of the five rating tokens, per attack, on (a) Themis and (b) Prometheus. Top row of each panel: Readability attacks; bottom row: Adequacy attacks. Crystallization (dashed line) occurs at $L\!=\!26$ on Themis and $L\!=\!25$ on Prometheus.}
\label{fig:logit-lens}
\end{figure*}

\subsection{Results and Analysis}


\paragraph{Errors register at early MLPs and route through mid-layer attention to the final position}

Early MLP layers locally register the perturbation at the perturbed-token position uniformly across attacks and across both evaluators. On Themis (Figure~\ref{fig:window-mode}a, top) and Prometheus (Figure~\ref{fig:window-mode}b, top), window-mode MLP causal effects concentrate at relative position $+0$ in layers 0-5 and are near zero elsewhere, with magnitudes strongest for the most token-localized attacks (Entity, Numerical, Antonym). The perturbed token's MLP output thus locally encodes the corrupted content irrespective of evaluation criterion.

Window-mode attention diverges in attack-specific ways that mirror what each comparison requires, with three patterns visible in both models (Figure~\ref{fig:window-mode}, bottom rows). Preposition spreads from $+0$ through $+2$-$3$ across L0-L10 (checking adjacent syntactic context); Entity and Numerical concentrate sharply at $+0$ in mid layers with no horizontal spread (retrieving from elsewhere in the article); and Coreference, Antonym, Tense, and Reordering produce weak window-mode attention overall, with their comparison happening through last-position attention to distant tokens outside our six-token window.

At the final position, all attacks share the same routing-and-integration architecture on Themis (Figure~\ref{fig:themis-last}): last-position attention $o_\text{proj}$ effects concentrate in a single band at L13-L15 where perturbation information is routed into the last token, after which the MLP effect rises in a stepped cascade with bands at L17-L18, L25-L27, and a peak at L31 that acts as the final write into the rating-token direction. Prometheus shares the same routing band but diverges at the top of the cascade (Figure~\ref{fig:prometheus-last}): its last-position attention spans more depths (L14-L17, L19-L21, L29-L30), and its L31 MLP effect is strongly negative where Themis's is strongly positive. Themis thus commits the rating in one dominant top-layer step; Prometheus distributes the commit across mid-to-late layers and uses L31 for an opposite-direction correction.

\paragraph{The rating decision crystallizes at a sharp late layer (L25–L26)}
The rating decision crystallizes at a sharp, late layer that is shared across all attacks. On Themis (Figure~\ref{fig:logit-lens}a), the logit-lens probability mass on all five rating tokens is essentially zero through $L \sim 22$ and then rises steeply over a narrow band, with every attack transitioning at the same depth; we refer to $L\!=\!26$ as the crystallization layer. Prometheus shows the same qualitative shape one layer earlier at $L\!=\!25$ (Figure~\ref{fig:logit-lens}b), indicating that depth-aligned late commitment is a property of both base architectures. Bootstrapping the max-slope depth across seeds yields a 95\% CI of [26, 26] on Themis and [25, 25] on Prometheus.

Post-crystallization, the model commits to different ratings for different attacks, revealing an implicit severity hierarchy. On Themis, Entity and Antonym crystallize to Rating 2 (harshest), Tense to Rating 4 (mildest), and the remaining attacks cluster at Rating 3. Prometheus reproduces this ordering more punitively (Entity and Antonym at Rating 1, most others at Rating 4) but with markedly less confident trajectories: where Themis's corrupt-rating probability climbs monotonically toward $\sim 1.0$ by the final layer, Prometheus's peaks at $\sim 0.5$ around $L\!=\!30$ and then declines, with a sustained Rating-5 prior visible pre-crystallization that is never fully suppressed. The same crystallization depth therefore underlies very different commitment trajectories, consistent with the divergent top-of-cascade MLP dynamics shown in Figures~\ref{fig:themis-last} and~\ref{fig:prometheus-last}.

\begin{figure*}[t]
  \centering
  \includegraphics[width=0.9\textwidth]{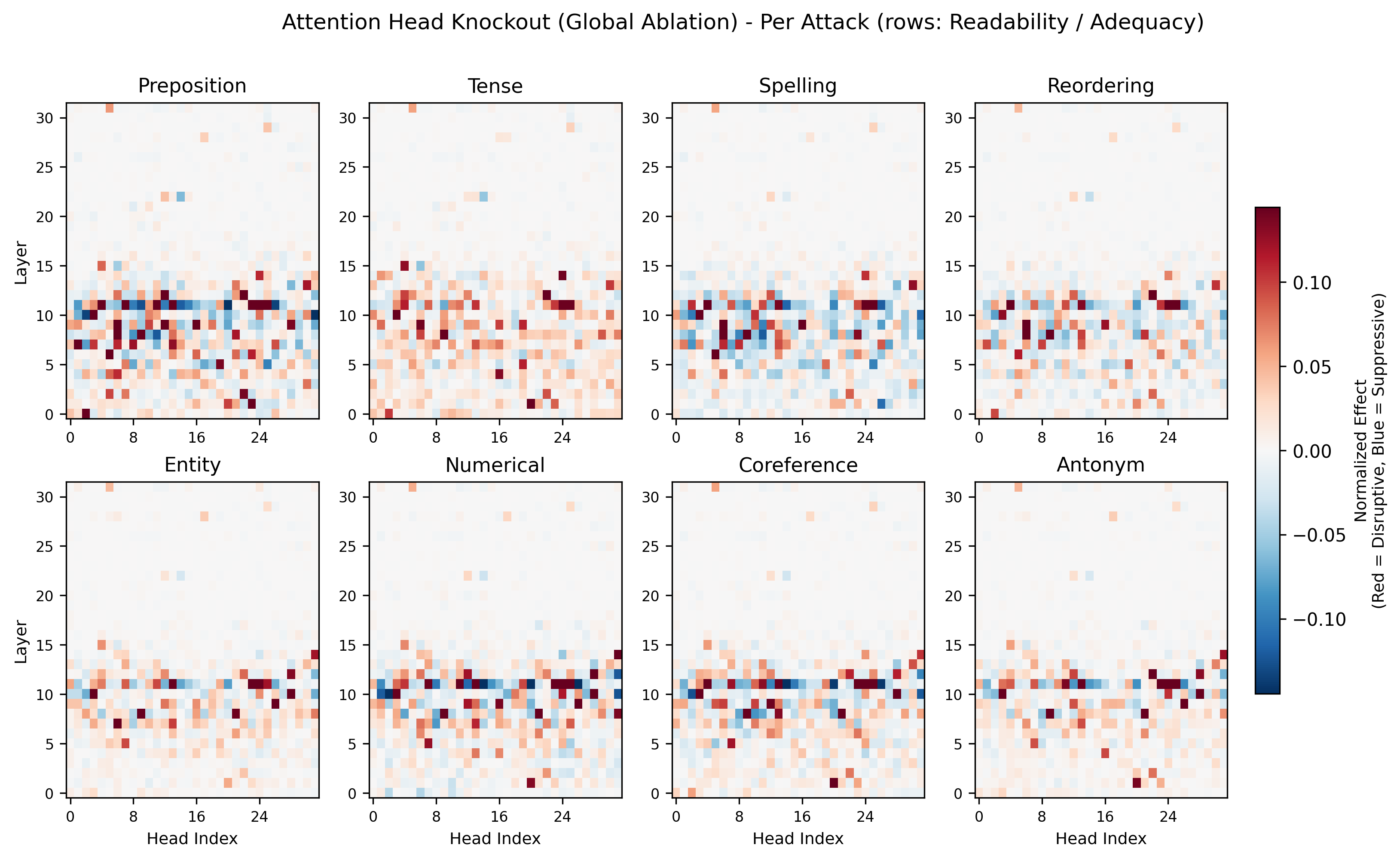}
  \caption{
Per-attack attention-head knockout heatmaps on Themis. Top row: Readability attacks (Preposition, Tense, Spelling, Reordering). Bottom row: Adequacy attacks (Entity, Numerical, Coreference, Antonym). The L10-L11 evaluation band appears in every individual attack panel, with both disruptive (red) and suppressive (blue) heads in an alternating pattern; Readability attacks additionally recruit heads at L7-L9, while Adequacy attacks remain sparse between L5 and L9.
}
\label{fig:themis-head-knockout-perattack}
\end{figure*}

\paragraph{A narrow band of attention heads at L10–L11 implements the rating verdict}

On Themis (Figure~\ref{fig:themis-head-knockout-perattack}), every individual attack panel shows a horizontal band at L10-L11 containing both disruptive (red) and suppressive (blue) heads in an alternating pattern, with negligible individual-head effects elsewhere. For each attack, we define an ablation-active head as one whose knockout produces a normalized causal effect exceeding a threshold at the last input 
position, and quantify concentration as the fraction of such heads falling 
within a specified layer range. L10-L11 contains $30.2\% \pm 0.9\%$ of 
Themis's ablation-active heads ($4.8\times$ enrichment over the uniform 
expectation from 32 layers), and L10-L15 contains $39.9\% \pm 0.5\%$ of 
Prometheus's. This alternating sign shows that the L10-L11 heads split into two functional roles, heads that push the rating toward the corrupt output, and heads that defend the clean rating and the final rating emerges from their interaction rather than from any single dominant head. This pattern holds regardless of attack type, indicating that a small, dedicated group of interacting heads implements the rating decision across error categories.

Within this shared band, the two criteria differ in upstream-head recruitment: the Readability panels (Figure~\ref{fig:themis-head-knockout-perattack}, top row) show additional head activity at L7-L9 contributing nearly as strongly as the L10-L11 band itself, while the Adequacy panels (bottom row) remain markedly sparser between L5 and L9. This matches the window-mode finding of Figure~\ref{fig:window-mode}: Readability's broader early-layer recruitment aligns with its local-attention spreading, while Adequacy's tighter L10-L11 profile aligns with the mid-layer concentration of Entity- and Numerical-type attention.

Prometheus exhibits the same evaluation-band concept but distributed more broadly across L10-L15, with the per-criterion sparsity pattern inverted relative to Themis (Readability denser, Adequacy sparser; Appendix Figure~\ref{fig:prometheus-head-knockout-perattack-app}). Both models show essentially no individual-head effects above L15, consistent with the division of labor identified across our experiments: attention does the comparison and routing work below L15, after which the upper-layer MLP cascade integrates the routed signal and writes the rating, with no individual head playing a critical role.

\paragraph{An End-to-End Two-Stage Evaluation Pipeline.}
These findings synthesize into a two-stage pipeline shared across both 
evaluators: below layer~15, attention performs comparison and routing 
(early MLPs encode the corrupted content at the perturbed token, an 
L10-L11 band of heads implements the rating verdict, and L13-L15 
attention routes the verdict to the final input position); above it, 
the MLP cascade integrates the routed signal and writes the rating, 
with the decision crystallizing at a sharp late layer ($L\!=\!26$ on 
Themis, $L\!=\!25$ on Prometheus). Prometheus shares this scaffold but diverges at the top of the cascade, using L31 for an opposite-direction correction where Themis writes in one dominant top-layer step. We show these two mechanisms on the Llama-3-8B substrate; whether Mistral-7B fine-tuning installs analogous or different mechanisms requires a matched base-model control on that architecture, which we leave to future work.

\paragraph{Finding generalization across data domains.}
To test whether the two-stage pipeline reflects a dataset-specific 
artifact of CNN/DM or a domain-general property of the evaluators, we 
further replicate the full causal-tracing and logit-lens sweep on a different dataset XSum \citep{narayan-etal-2018-dont},a benchmark of BBC news articles paired with 
single-sentence, highly abstractive summaries. XSum's shorter, more 
abstractive style contrasts with CNN/DM's largely-extractive multi-sentence bullets, making it a materially different distribution for 
testing domain transfer (three seeds, all eight attacks; see 
Table~\ref{app:xsum-generalization}). The crystallization depth is 
preserved exactly across both models, Themis at $L\!=\!26$ 
(95\% bootstrap CI $[26, 26]$, $N=2,242$ pooled samples) and Prometheus 
at $L\!=\!25$ ($[25, 25]$, $N=2,829$), with concentration ranging from 
89\% to 100\% of samples within $\pm 1$ layer of the median. The two-stage 
MLP separation replicates on Themis with the same qualitative structure 
(mean ratio $21.9\times$ across 7 attacks with defined ratios; 
below-L15 $= 0.007$, above-L15 $= 0.153$). On Prometheus, the below-L15 
MLP contribution becomes slightly \emph{negative} across all attacks 
($-0.005$ to $-0.001$) while the above L15 magnitude is preserved 
($\approx 0.09$), indicating that the more abstractive summaries in XSum compared to CNN/DM sharpen rather than weaken stage separation in this model. 
The 
routing band replicates on Themis with attention peaks at L13 across 
all attacks, while Prometheus's attention peak shifts one layer up to 
L16, broadening the routing band to L13-L16 on this model. The L31 
divergence between the two evaluators is preserved: Themis writes 
positively (+0.26 to +0.50 across attacks), Prometheus writes 
negatively (-0.07 to -0.15). The pipeline structure - routing 
followed by upper-cascade MLP integration with sharp late crystallization, is therefore preserved across a substantially more abstractive 
domain, though specific quantitative anchors (Prometheus's routing peak) 
shift slightly with the domain.

\paragraph{What fine-tuning installs: a base-model control.}
To isolate the specific effects of judge fine-tuning, we run the full 
causal-tracing sweep on the un-fine-tuned base model (Llama-3-8B) from 
which Themis is derived, using the same eight-attack taxonomy, 
single-instruction prompt, and CNN/DM samples (three seeds; see 
Table~\ref{app:base-model-completion}). The base model reproduces the 
routing architecture, the L13 attention band and the L15 MLP peak, and 
its above L15 MLP magnitude matches Themis's ($0.131$ vs.\ $0.125$, mean 
across attacks $\times$ 3 seeds), indicating that the scale of 
rating-writing in the upper cascade is set by the architecture rather 
than by fine-tuning. Two mechanisms differ sharply. First, the two-stage 
separation is much weaker on the base model: below-L15 MLP effect is 
$0.056$ versus Themis's $0.010$, a $\sim$5.5$\times$ gap that fine-tuning 
closes by suppressing below-L15 contribution at the last position. 
Second, the crystallization depth advances by two layers, from $L\!=\!28$ 
(95\% bootstrap CI $[28, 28]$, N=1{,}927 samples over 3 seeds) to Themis's 
$L\!=\!26$. Together, these two effects identify what judge fine-tuning 
installs on this substrate: below-L15 MLP suppression (producing the 
stage-1/stage-2 separation) and a two-layer compression of the rating 
cascade. Base-model behavioral sensitivity is also lower (about 39\% 
of pairs produce a rating change under single-instruction prompting, 
pooled across attacks, versus much higher for Themis).

\section{Conclusion}
\label{sec:conclusion}

Across four mechanistic experiments on two LLM-based evaluators - Themis (Llama-3-8B) and Prometheus (Mistral-7B), we find that NLG judges implement a structured, coherent evaluation pipeline rather than an ad-hoc rating-generation shortcut: attention performs local error comparison and routes the result to the final position below L15, after which the upper-layer MLP cascade integrates the routed signal and writes the rating, with the decision crystallizing in the residual stream at a sharp late layer ($L\!=\!26$ on Themis, $L\!=\!25$ on Prometheus). This pipeline structure and the associated crystallization depths consistently emerge across data domains, suggesting that these mechanisms capture a general computational strategy underlying LLM-based summarization evaluation. A base-model control at the same scale reproduces the routing architecture and the sharp crystallization property but not the stage separation, showing that judge fine-tuning installs two specific, localizable mechanisms, suppression of below-L15 MLP contribution and a two-layer advance of the crystallization depth, on top of a general transformer substrate rather than assembling the evaluation pipeline from scratch. This decomposition suggests that targeted interventions at these two loci may reproduce judge behavior without full task-specific fine-tuning. We release the eight-attack taxonomy, generation pipeline, and behaviorally verified clean/corrupt corpus as public resources to support further interpretability research on NLG evaluators and downstream applications such as contrastive evaluator training and localized-evaluator distillation.
\section{Limitations}
\label{sec:limitations}

\paragraph{Model and task scope.} Our circuit-level findings are based on two open-source evaluators (Themis built on Llama-3-8B; Prometheus built on Mistral-7B) applied to English-language summaries drawn from a single source distribution (CNN/DailyMail), using a fixed evaluation prompt per evaluator (Appendix~\ref{app:prompts}). While our taxonomy is designed to be task-agnostic and the two base architectures are drawn from distinct model families, we have not directly verified the identified circuit on dedicated evaluators of dialogue response generation, story generation, or factuality assessment outside summarization, nor on larger or multilingual evaluator models. Generalization of the two-stage circuit structure to these settings, and its stability under prompt variation, remains an empirical question.

\paragraph{Behavioral-filter selection.} As described in §\ref{sec:taxonomy}, our analysis is restricted to samples whose evaluator rating demonstrably changes under perturbation. This focuses our circuit characterization on cases where the evaluation procedure successfully detects the perturbation; samples on which the evaluator misses the perturbation entirely fall outside our analysis. Such failure-mode samples carry interpretable signal about evaluator robustness that the present work does not address.

\paragraph{Single intensity and single-attack samples.} Single intensity and single-attack samples. All experiments use single-attack samples at intensity k = 1. Behavior under higher intensities or under mixed-attack samples (two or more perturbation categories co-occurring) is not characterized here; we flag this as a follow-up direction.

\paragraph{Cross-family claims rest on two evaluators.} Cross-family claims rest on two evaluators. Patterns described as 'shared' or 'family-specific' are supported by n = 2 evaluators, one per architecture. These should be read as evidence from two independent data points rather than as generalizations across full model families or the broader space of LLM-based NLG evaluators.

\paragraph{Logit-lens approximation.} Our depth-resolved decoding of rating commitments uses the logit lens, which approximates intermediate beliefs by projecting the residual stream through the final unembedding. Where the residual stream is rotated or scaled relative to its final-layer geometry, the logit lens can under-report probability mass on the eventual rating; tuned-lens variants address this but introduce per-layer training that is not directly comparable across base-model families. We chose the logit lens for cross-architecture consistency and treat absolute pre-crystallization probabilities as qualitative rather than calibrated.

\paragraph{Generator-model dependence.} Generator-model dependence. Perturbations are produced by GPT-4o. Any systematic stylistic bias could correlate with evaluator response patterns in ways this work does not fully isolate. We mitigate through behaviorally-verified pairs and explicit position tracking, but a generator-independent characterization is outside the present scope.

\section{Ethics Statement}
\label{sec:ethics}
All experiments used publicly available datasets (CNN/DailyMail, 
\citealp{hermann2015teaching}) and publicly released evaluator models 
(Themis, \citealp{hu2024themis}; Prometheus, \citealp{kim2024prometheus}). 
No personally identifiable information was collected, and no human 
subjects were involved beyond the datasets' original construction. 
Perturbation generation uses GPT-4o via standard API access; generated 
perturbations are limited to controlled syntactic and semantic alterations 
of public news summaries.

We release the perturbation taxonomy, generation pipeline, and 
behaviorally-verified corpus as public resources for further mechanistic 
study of NLG evaluators. These artifacts could in principle be repurposed 
for adversarial attacks, but we view this risk as limited: the perturbation 
categories we study are well-established in the prior NLG-evaluation 
literature, and our findings do not by themselves provide novel attack 
vectors against deployed LLM-judges. We assess the net contribution as 
positive: better evaluator development through mechanistic understanding 
of how current evaluators operate.

\paragraph{Use of AI Assistants.} We used AI assistants (Claude and 
ChatGPT) during the preparation of this work for two purposes: (i) 
supporting code implementation and debugging of our causal tracing and 
attention-head-knockout infrastructure, and (ii) language polishing, 
proofreading, and improving the stylistic clarity of the manuscript. 
AI assistants were not used to generate research questions, design 
experiments, or interpret results. All research design, experimental 
decisions, scientific interpretations, and final text were developed 
and verified by the human authors, who hold full responsibility for 
the integrity and accuracy of this work.

\section{Acknowledgments}
\label{sec:acknowledgements}

We sincerely appreciate the valuable feedback provided by our reviewers, which greatly helped to improve the manuscript. This research is partially supported by the National Science Foundation (IIS-2604291). The content is solely the responsibility of the authors and does not necessarily represent the official views of the National Science Foundation.

\bibliography{emnlp26_ref}

\appendix

\section{Evaluation Prompts}
\label{app:prompts}

\subsection{Themis Prompt}
\label{app:prompts-themis}

The Themis evaluator uses the single-instruction prompt shown in Figure~\ref{fig:themis-prompt}.

\begin{figure*}[t]
\begin{Verbatim}[
  frame=single,
  framesep=6pt,
  rulecolor=\color{gray!60},
  fontsize=\small,
  baselinestretch=1.0
]
###Instruction###
Please act as an impartial and helpful evaluator for natural language generation (NLG).
Your task is to evaluate the quality of the Summarization strictly based on the 
given evaluation criterion.
You MUST keep to the strict boundaries of the evaluation criterion.
**Scoring Rules:**
1. Use a Likert scale from 1 to 5.
2. 5 is the highest quality, 1 is the lowest.
3. Output ONLY the number.
###Evaluation Criterion###
{prompt_criteria}
###Task###
Article:
{article}
Summary:
{summary}
###Rating###
The rating is
\end{Verbatim}
\caption{Themis prompt template. Braced fields are Python format-string placeholders substituted at runtime.}
\label{fig:themis-prompt}
\end{figure*}

\subsection{Prometheus Prompt}
\label{app:prompts-prometheus}

Prometheus requires a few-shot variant of the prompt to reliably emit the rating token at the first generation position. The prompt template is shown in Figure~\ref{fig:prometheus-prompt}. The \texttt{\{examples\_block\}} placeholder is filled with the five article/summary/rubric/rating exemplars listed below, which span the full $1$--$5$ rating range. All five share the same scoring rubric: ``Evaluate the overall quality of the summary based on how accurately and completely it captures the main facts of the article. Rate from 1 (highly inaccurate or off-topic) to 5 (fully accurate and comprehensive).''

\begin{figure*}[t]
\begin{Verbatim}[
  frame=single,
  framesep=6pt,
  rulecolor=\color{gray!60},
  fontsize=\small,
  baselinestretch=1.0
]
###Task Description:
You are an impartial evaluator. Read the article, the response, and the score rubric. 
Output ONLY a single integer between 1 and 5 representing your rating. 
Do not write any feedback, explanation, or any other text, just the digit. 
The examples below illustrate the expected output format.
{examples_block}
###Now evaluate this:
###Article:
{article}
###Response to evaluate:
{summary}
###Score Rubric:
{prompt_criteria}
\end{Verbatim}
\caption{Prometheus prompt template. Braced fields are Python format-string placeholders substituted at runtime.}
\label{fig:prometheus-prompt}
\end{figure*}

This appendix lists the full prompt templates used to elicit ratings from each evaluator. Braced fields (\texttt{\{article\}}, \texttt{\{summary\}}, \texttt{\{prompt\_criteria\}}, \texttt{\{examples\_block\}}) are Python format-string placeholders substituted at runtime.

\paragraph{Example 1 (Rating 4).} \textit{Article:} The Federal Reserve announced a 0.25 percentage point interest rate cut today, the third such reduction this year, citing concerns about slowing economic growth and persistent global trade tensions affecting U.S.\ manufacturing output. \textit{Summary:} The Fed cut interest rates by 0.25 points today due to economic growth concerns.

\paragraph{Example 2 (Rating 2).} \textit{Article:} The European Union announced a new trade agreement with Vietnam yesterday, eliminating tariffs on 99 percent of goods traded between the two regions over the next seven years and establishing new standards for labor and environmental protections. \textit{Summary:} Vietnam and the EU had a meeting about trade and economic cooperation last week.

\paragraph{Example 3 (Rating 5).} \textit{Article:} Scientists at Stanford University have developed a new lithium-metal battery technology that can charge a smartphone in 30 seconds and last three times longer than current lithium-ion batteries, with potential applications in electric vehicles and grid storage. \textit{Summary:} Stanford researchers developed a lithium-metal battery that charges phones in 30 seconds and lasts three times longer than lithium-ion, with potential uses in electric vehicles and grid storage.

\paragraph{Example 4 (Rating 1).} \textit{Article:} NASA's Perseverance rover has identified organic compounds in rock samples from Mars's Jezero crater, providing the strongest evidence yet of conditions that could have supported ancient microbial life on the planet, according to a study published in \emph{Nature}. \textit{Summary:} Astronauts from China landed on the moon last week to collect lunar samples for future research.

\paragraph{Example 5 (Rating 3).} \textit{Article:} A magnitude 6.2 earthquake struck off the coast of northeastern Japan early Tuesday morning, causing minor structural damage to buildings in three coastal cities and triggering brief tsunami advisories that were lifted within two hours. No casualties have been reported. \textit{Summary:} An earthquake hit Japan and caused some damage. There were no injuries.

\begin{table*}[t]
\centering
\small
\begin{tabular}{lcccc}
\toprule
\textbf{Attack} & \textbf{Below-L15} & \textbf{Above-L15} & \textbf{Ratio} & \textbf{Cryst. depth} \\
 & (mean $\pm$ SEM) & (mean $\pm$ SEM) & & (95\% CI) \\
\midrule
\multicolumn{5}{l}{\textit{Adequacy attacks}} \\
Entity          & $0.010 \pm 0.001$ & $0.125 \pm 0.010$ & $12.77\times$ & L26 [26, 26] \\
Numerical       & $0.012 \pm 0.002$ & $0.143 \pm 0.014$ & $12.41\times$ & L26 [26, 26] \\
Coreference     & $0.010 \pm 0.001$ & $0.158 \pm 0.008$ & $16.16\times$ & L26 [26, 26] \\
Polarity        & $0.008 \pm 0.001$ & $0.127 \pm 0.010$ & $15.10\times$ & L26 [26, 26] \\
\midrule
\multicolumn{5}{l}{\textit{Readability attacks}} \\
Preposition     & $0.008 \pm 0.003$ & $0.176 \pm 0.006$ & $20.98\times$ & L26 [26, 26] \\
Tense           & $0.013 \pm 0.004$ & $0.177 \pm 0.008$ & $14.19\times$ & L26 [26, 26] \\
Spelling        & $0.010 \pm 0.002$ & $0.165 \pm 0.004$ & $15.76\times$ & L26 [26, 26] \\
Reordering      & $0.007 \pm 0.000$ & $0.137 \pm 0.008$ & $20.13\times$ & L26 [26, 26] \\
\midrule
\textbf{Mean}   & $\mathbf{0.010}$  & $\mathbf{0.151}$  & $\mathbf{15.56\times}$ & \textbf{L26 [26, 26]} \\
\bottomrule
\end{tabular}
\caption{Themis (Llama-3-8B) causal tracing under single-instruction prompting, three seeds \{17, 23, 42\}, on CNN/DM. Two-stage split reports the mean last-position MLP causal effect below L15 and at/above L15 (pooled across attacks × 3 seeds), with the above/below ratio. Crystallization depth is the max-slope layer from logit-lens 5-rating probability trajectories, pooled across seeds. Mean row aggregates all 8 attacks $\times$ 3 seeds ($N=1361$ pooled).}
\label{tab:themis-full}
\end{table*}

\begin{figure*}[t]
  \centering
  \includegraphics[width=\textwidth]{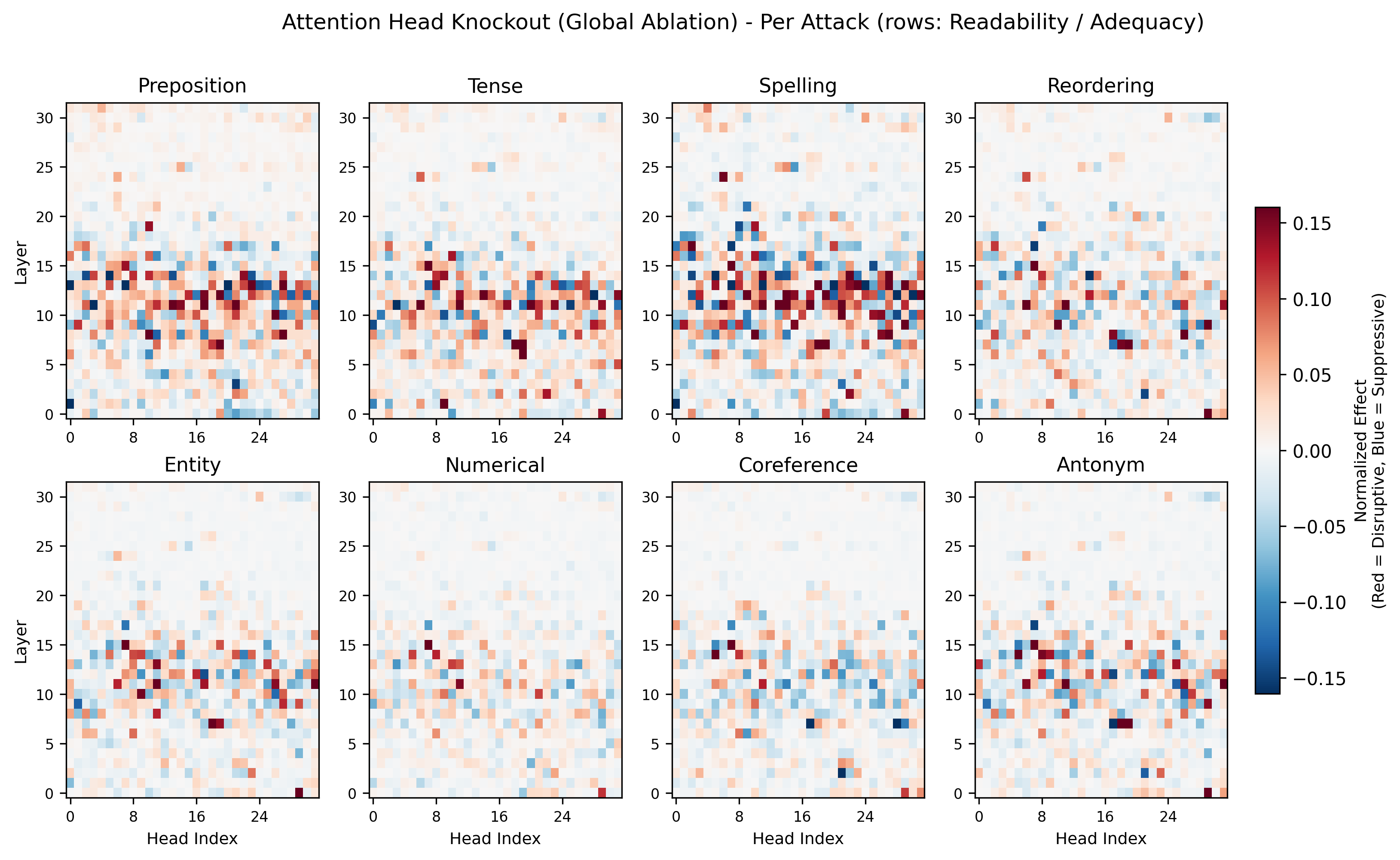}
  \caption{
Per-attack attention-head knockout heatmaps on Prometheus. Top row: Readability attacks; bottom row: Adequacy attacks. The broader L10-L15 evaluation band is present in individual attack panels, with the Readability panels showing denser head activity than the Adequacy panels, the inverse of the per-criterion sparsity pattern observed on Themis (Figure~\ref{fig:themis-head-knockout-perattack}).
}
\label{fig:prometheus-head-knockout-perattack-app}
\end{figure*}

\begin{table*}[t]
\centering
\small
\begin{tabular}{lcccccc}
\toprule
& \multicolumn{3}{c}{\textbf{Themis (Llama-3-8B)}} & \multicolumn{3}{c}{\textbf{Prometheus (Mistral-7B)}} \\
\cmidrule(lr){2-4} \cmidrule(lr){5-7}
\textbf{Attack} & \textbf{Below-L15} & \textbf{Above-L15} & \textbf{Cryst.} & \textbf{Below-L15} & \textbf{Above-L15} & \textbf{Cryst.} \\
\midrule
\multicolumn{7}{l}{\textit{Adequacy attacks}} \\
Entity      & $0.008$ & $0.111$ & L26 & $-0.003$ & $0.083$ & L25 \\
Numerical   & $0.007$ & $0.161$ & L26 & $-0.001$ & $0.085$ & L25 \\
Coreference & $0.006$ & $0.189$ & L26 & $-0.003$ & $0.089$ & L25 \\
Polarity    & $0.006$ & $0.131$ & L26 & $-0.002$ & $0.076$ & L25 \\
\midrule
\multicolumn{7}{l}{\textit{Readability attacks}} \\
Preposition & $0.008$ & $0.155$ & L26 & $-0.004$ & $0.095$ & L25 \\
Tense       & $0.008$ & $0.184$ & L26 & $-0.005$ & $0.091$ & L25 \\
Spelling    & $-0.005$ & $0.143$ & L26 & $-0.003$ & $0.091$ & L25 \\
Reordering  & $0.006$ & $0.141$ & L26 & $-0.005$ & $0.093$ & L25 \\
\midrule
\textbf{Mean} & $\mathbf{0.007}^{\dagger}$ & $\mathbf{0.153}^{\dagger}$ & \textbf{L26 [26, 26]} & $\mathbf{-0.003}$ & $\mathbf{0.088}$ & \textbf{L25 [25, 25]} \\
\bottomrule
\end{tabular}
\caption{XSum causal tracing and crystallization for both evaluators (three seeds \{17, 23, 42\}; single-instruction prompting for Themis, few-shot for Prometheus). Two-stage split reports mean last-position MLP causal effect below and at/above L15. Crystallization depth is the max-slope layer from logit-lens 5-rating probability trajectories, with 95\% bootstrap CI over pooled seeds shown for the mean row. On Themis the below/above ratio is $\sim 22\times$; on Prometheus the below-L15 contribution is slightly negative, indicating pure stage-2 rating writing. Total pooled samples: $N=2{,}242$ for Themis, $N=2{,}829$ for Prometheus.
$^{\dagger}$Themis mean excludes Spelling (below-L15 is slightly negative, ratio undefined); above-L15 mean is over all 8 attacks.}
\label{app:xsum-generalization}
\end{table*}

\section{Window-Mode Causal Tracing: Methodology Details}
\label{app:tracing-details}

This appendix documents the filtering, alignment, and aggregation procedures used for window-mode causal tracing (§\ref{sec:methods}). Last-token-mode tracing and head knockout follow the standard logit-difference formulation and do not require these additional controls, as their per-sample yield is not constrained by per-token alignment.

\paragraph{Sample-level filters.} Three filters are applied before per-cell aggregation. (i) \emph{Non-rating output filter:} samples whose clean forward pass does not assign substantial probability mass to the rating-token set are discarded; this is rare but catches degenerate prompts. (ii) \emph{Score-difference filter:} we require $|E_\text{clean} - E_\text{corrupt}| \geq 0.05$, corresponding to approximately $5\%$ probability mass shifting between adjacent integer ratings. Samples below this threshold yield uninterpretable patching effects because the denominator of Equation~1 approaches zero. Pass rates under this filter are roughly $25$--$40\%$ for Readability attacks and $50$--$70\%$ for Adequacy attacks, themselves an indicator of evaluator sensitivity per attack. (iii) \emph{Anchor-validity filter:} per-token edit operations between clean and corrupt sequences are identified using \texttt{difflib} sequence-matching opcodes (\texttt{replace}, \texttt{insert}, \texttt{delete}, \texttt{equal}), and cross-referenced against the perturbation pipeline's recorded modification positions to confirm the anchor is at the intended site. For Reordering, where the pipeline's recorded swap can be inaccurate relative to the emitted text, we use difflib directly.

\paragraph{Token-length mismatch handling.} Several perturbation categories (notably Spelling, Numerical, and Reordering) produce corrupt sequences of different length than clean: BPE tokenizers split misspellings and numerical substrings inconsistently. Naïvely indexing the clean cache at the corrupt-side position is therefore incorrect. We compute a per-token clean-source-position mapping from the difflib opcodes: \texttt{equal} opcodes yield a one-to-one mapping (\texttt{clean\_pos = corrupt\_pos + offset}); \texttt{replace} opcodes are clamped to the last valid clean position within the replaced span; and \texttt{insert} opcodes have no clean correspondence and are recorded as \texttt{NaN}. When patching at (layer, target position) in the corrupt run, we read the source activation from (layer, mapped clean-source position) in the clean run; cells with no valid alignment remain \texttt{NaN} and are excluded from the per-cell aggregate.

\paragraph{Per-cell aggregation.} For each (layer, relative-position) cell, we accumulate effect values from all samples retained after filtering, pooled across three random seeds ($23$, $42$, $17$) each sampling a different 200-sample subset of the source corpus. The number of contributing samples varies per cell and per attack: well-behaved attacks have $N \approx 400$--$500$ per cell after seed pooling, while attacks with BPE-unstable perturbations retain as few as $\sim 20$ per cell (notably Themis on Spelling, where the Llama-3 tokenizer splits misspellings inconsistently). We aggregate per cell with a $20\%$ trimmed mean: dropping the smallest and largest $10\%$ of contributions and averaging the rest. This is the standard robust-statistic choice; plain mean is outlier-dominated at small $N$, while the median over-collapses the heavy-tailed distributions we observe (the median sits in a near-zero bulk and discards genuine signal). Cells with $N < 10$ are additionally masked as \texttt{NaN} and render in light gray.

\paragraph{Color scaling.} Effect magnitudes differ by roughly an order of magnitude between Readability and Adequacy attacks (and between MLP and attention rows). To preserve within-criterion visibility while keeping the cross-criterion magnitude difference itself legible, each figure uses four independent color scales (one per row $\times$ criterion combination), displayed via four colorbars: two between the criterion groups and two on the right edge. The colormap is symmetric (\texttt{RdBu}) around zero.

\begin{table*}[t]
\centering
\small
\begin{tabular}{lcccc}
\toprule
\textbf{Attack} & \textbf{Below-L15} & \textbf{Above-L15} & \textbf{Ratio} & \textbf{Cryst. depth} \\
 & (mean $\pm$ SEM) & (mean $\pm$ SEM) & & (95\% CI) \\
\midrule
\multicolumn{5}{l}{\textit{Adequacy attacks}} \\
Entity          & $0.053 \pm 0.021$ & $0.131 \pm 0.004$ & $2.48\times$ & L28 [28, 28] \\
Numerical       & $0.052 \pm 0.014$ & $0.126 \pm 0.005$ & $2.42\times$ & L28 [28, 28] \\
Coreference     & $0.045 \pm 0.006$ & $0.121 \pm 0.004$ & $2.69\times$ & L28 [28, 28] \\
Polarity        & $0.046 \pm 0.014$ & $0.129 \pm 0.003$ & $2.84\times$ & L28 [28, 28] \\
\midrule
\multicolumn{5}{l}{\textit{Readability attacks}} \\
Preposition     & $0.064 \pm 0.017$ & $0.136 \pm 0.006$ & $2.13\times$ & L28 [28, 28] \\
Tense           & $0.065 \pm 0.020$ & $0.137 \pm 0.008$ & $2.12\times$ & L28 [28, 28] \\
Reordering      & $0.068 \pm 0.027$ & $0.138 \pm 0.004$ & $2.02\times$ & L28 [28, 28] \\
Spelling        & \multicolumn{3}{c}{--- see footnote$^{\dag}$ ---} & L28 [28, 28] \\
\midrule
\textbf{Mean}   & $\mathbf{0.056}$  & $\mathbf{0.131}$  & $\mathbf{2.34\times}$ & \textbf{L28 [28, 28]} \\
\midrule
\textit{Themis (reference)} & $0.010$ & $0.125$ & $12.8\times$ & L26 [26, 26] \\
\bottomrule
\end{tabular}
\caption{Base-model (Llama-3-8B) causal tracing under single-instruction (completion) prompting. Two-stage separation is weak (mean 2.34$\times$) and above-L15 MLP magnitude matches Themis, while below-L15 MLP suppression is absent; fine-tuning drives below-L15 from 0.056 to 0.010 (~5.5$\times$ reduction) while preserving the above L15 cascade magnitude. Crystallization depth is L28 across all attacks (N=1,927 pooled samples), two layers above Themis's L26. $^{\dag}$Spelling causal tracing yielded no valid samples after the anchor-validity filter due to BPE tokenizer instability on misspelled tokens; the same limitation is documented for Themis Spelling on CNN/DM in Appendix~\ref{app:tracing-details}.}
\label{app:base-model-completion}
\end{table*}

\subsection{Verification Protocol for Injected Perturbations}
\label{app:verification}

At every generation run we manually inspect 20 clean/corrupt 
pairs per attack against three criteria:
(i) that a perturbation was actually injected, GPT-4o occasionally returns 
the summary unmodified when a short summary offers no valid target (e.g., 
no swappable preposition);
(ii) that the returned \texttt{original\_tokens}/\texttt{new\_tokens} maps 
correctly identify the changed span, verified against the clean and 
perturbed summaries. This is the critical check because the token map 
anchors every causal intervention, and an incorrect anchor would 
invalidate the trace;
(iii) that the perturbation matches its declared type e.g., a Spelling 
attack introduces an orthographic error rather than a preposition or 
tense change.

Typically 19 to 20 of 20 pairs pass all three criteria. Failures are 
overwhelmingly of type (i) and are removed automatically by a structural 
filter (empty \texttt{new\_tokens}) before analysis. We additionally read 
summaries for naturalness: the corrupted outputs read as fluent, plausible 
model output rather than artificially mangled text.

\paragraph{Cross-attack severity.} 
Equal severity across attack types is not a property our design targets. 
Prior work whose aspect-targeted attacks are the closest antecedent to our 
taxonomy shows that different attack types produce systematically different 
effects on evaluator judgments \citep{hu2024llm}. This is expected: 
a human evaluator would not penalize a spelling error as heavily as an 
entity swap that makes the summary factually wrong. Our eight attacks 
span two quality axes to observe how the evaluator's internal computation 
differs as a function of error type and quality criterion, not to build a 
severity-matched stimulus set. Every attack is matched on the dimension 
our method requires, perturbation intensity ($k=1$), which makes the 
causal traces comparable across attacks; severity is the variable under 
study. A formal human annotation of severity is left to 
future work.



\section{Model Size, Computational Budget, and Infrastructure}
\label{app:compute}

\paragraph{Model parameters.}
Our analysis targets two open-source NLG evaluator models: 
\textbf{Themis}~\citep{hu2024themis}, built on Llama-3-8B with approximately 
$8$~billion parameters, and \textbf{Prometheus-7B-v2.0}~\citep{kim2024prometheus}, 
built on Mistral-7B with approximately $7$~billion parameters. Both 
models share a comparable transformer architecture: $32$ layers, 
$32$ attention heads per layer, and a hidden dimension of $4096$. 
Perturbation generation uses \textbf{GPT-4o} via the OpenAI API; 
its parameter count is not publicly disclosed by the provider.

\paragraph{Computing infrastructure.}
The majority of mechanistic-analysis experiments (approximately $80\%$ of 
GPU-hours) were conducted on a cluster of $16 \times$ \textbf{NVIDIA GeForce 
RTX 2080 Ti} GPUs ($11$~GB VRAM each), with the remaining $20\%$ run on a 
single \textbf{NVIDIA H100} ($80$~GB) GPU. Models were loaded in 
\textbf{bfloat16} for inference and activation patching, with model-parallel 
sharding across the eight 2080 Ti devices to accommodate the $7$--$8$~B 
parameter evaluators. Implementation used PyTorch and the HuggingFace 
\texttt{transformers} library.

\paragraph{Computational budget.}
The full experimental battery used approximately \textbf{800} 
GPU-hours, distributed as follows:
\begin{itemize}
    \itemsep0pt
    \item \textbf{Perturbation generation} (GPT-4o API): approximately 
    \textbf{5000} API calls across $8$ attacks $\times$ $200$ samples 
    $\times$ $3$ seeds, at an approximate cost of \textbf{\$100}.
    \item \textbf{Window-mode and last-token-mode causal tracing}: 
    \textbf{250} GPU-hours, dominated by per-layer/per-position activation 
    patching across $32$ layers $\times$ both MLP and attention sublayers.
    \item \textbf{Logit-lens analysis}: \textbf{150} GPU-hours.
    \item \textbf{Attention-head knockout} ($32 \times 32 = 1024$ heads 
    per model, both evaluators): \textbf{400} GPU-hours.
\end{itemize}
We estimate the full experimental battery is reproducible with approximately 
$800$ GPU-hours on equivalent consumer-grade hardware.

\end{document}